\documentclass[10pt,twocolumn,letterpaper]{article}
\usepackage[utf8]{inputenc}

\usepackage{cvpr}
\usepackage{times}
\usepackage{epsfig}
\usepackage{graphicx}
\usepackage{amsmath}
\usepackage{amssymb}
\usepackage{bm}

\usepackage{bm}
\usepackage{bbm}

\newcounter{nbdrafts}
\makeatletter
\newcommand{\checknbdrafts}{
\ifnum \thenbdrafts > 0
\@latex@warning@no@line{**********************************************************************}
\@latex@warning@no@line{* The document contains \thenbdrafts \space draft note(s)}
\@latex@warning@no@line{**********************************************************************}
\fi}

\makeatother

\newcommand{\comment}[1]{}

\newcommand{\bF}[0]{\mathbf{F}}
\newcommand{\bP}[0]{\mathbf{P}}

\newcommand{\bB}[0]{\mathbf{B}}

\newcommand{\bX}[0]{\mathbf{X}}

\newcommand{\bI}[0]{\mathbf{I}}

\newcommand{\bb}[0]{\mathbf{b}}

\newcommand{\balpha}[0]{\bm{\alpha}}
\newcommand{\bmu}[0]{\bm{\mu}}
\newcommand{\feta}[0]{\bm{\eta}}
\newcommand{\bh}[0]{\bm{h}}
\newcommand{\be}[0]{\bm{e}}
\newcommand{\bp}[0]{\bm{p}}
\newcommand{\ff}[0]{\bm{f}}

\newcommand{\mL}[0]{\mathcal{L}}

\newcommand{\mR}[0]{\mathbb{R}}
\newcommand{\mI}[0]{\mathcal{I}}
\newcommand{\mN}[0]{\mathcal{N}}
\newcommand{\mT}[0]{\mathcal{T}}

\newcommand{\ONE}[0]{\mathbbm{1}}
\DeclareMathOperator*{\argmin}{arg\,min}

\newcommand{\smttt}[1]{{\small{\texttt{#1}}}}

\usepackage[breaklinks=true,bookmarks=false]{hyperref}

\cvprfinalcopy 

\begin{document}


\title{Social Scene Understanding:\\ End-to-End Multi-Person Action Localization and Collective Activity Recognition}

\author{Timur Bagautdinov$^1$, 
Alexandre Alahi$^2$, 
François Fleuret$^{1,3}$, 
Pascal Fua$^1$, 
Silvio Savarese$^2$\\
$^1$École Polytechnique Fédérale de Lausanne (EPFL)\\
$^2$Stanford University\\
$^3$IDIAP Research Institute\\
{\tt\small \{timur.bagautdinov, francois.fleuret, pascal.fua\}@epfl.ch, 
{\tt\small \{alahi, ssilvio\}@stanford.edu}}
}

\maketitle

\begin{abstract}
We present a unified framework for understanding human social behaviors in raw image
sequences. 
Our model jointly detects multiple individuals, infers their social actions, and 
estimates the collective actions with a single feed-forward pass through a neural network. 
We propose a single architecture that does not rely on external detection 
algorithms but rather is trained end-to-end to generate
dense proposal maps that are refined via a novel inference scheme.
The temporal consistency is handled via a person-level matching Recurrent Neural
Network. The complete model takes as input a sequence of frames and outputs
detections along with the estimates of individual actions and collective activities. We
demonstrate state-of-the-art performance of our algorithm on multiple publicly
available benchmarks.
\end{abstract}

\vspace{-0.5cm}
\section{Introduction}
\vspace{-0.15cm}

\begin{figure}[ht!]
\begin{center}
\includegraphics[width=0.45\textwidth]{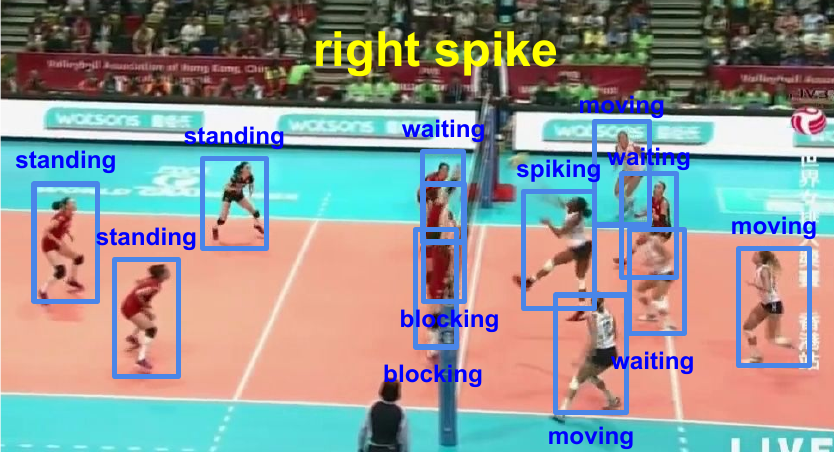}
\end{center}
\caption{Jointly reasoning on social scenes. Our method takes as input raw image
sequences and produces a comprehensive social scene interpretation: locations of
individuals (as bounding boxes), their individual social actions (e.g., ``blocking"), and the collective activity (``right spike" in the illustrated example).}
\label{fi:intro:pull}
\vspace{-0.5cm}
\end{figure}

Human social behavior can be characterized by ``social actions'' -- an
individual act which nevertheless takes into account the behaviour of other
individuals -- and ``collective actions'' taken together by a group of people
with a common objective. For a machine to perceive both of these actions, it
needs to develop a notion of collective intelligence, \textit{i.e.}, reason
jointly about the behaviour of multiple individuals. In this work, we propose a
method to tackle such intelligence. Given a sequence of image frames, our method 
jointly locates and describes the
social actions of each individual in a scene as well as the collective actions (see
Figure \ref{fi:intro:pull}). This perceived social scene representation can be
used for sports analytics, understanding social behaviour, surveillance, and
social robot navigation.

Recent methods for multi-person scene understanding take a sequential approach
\cite{Ibrahim2016,Deng2016,Ramanathan2016}: i) each person is detected in
every given frame; ii) these detections are associated over time by a
tracking algorithm; iii) a feature representation is extracted for each
individual detection; and finally iv) these representations are joined via
a structured model. Whereas the aforementioned pipeline seems reasonable, it
has several important drawbacks.  
First of all, the vast majority of state-of-the-art detection methods do not use any
kind of joint optimization to handle multiple objects, but rather rely on heuristic post-processing, and thus are susceptible to greedy non-optimal decisions. Second,
extracting features individually for each object discards a large amount of context and
interactions, which can be useful when reasoning about collective behaviours. 
This point is particularly important because the locations and actions of humans can be
highly correlated. 
For instance, in team sports, the location and action of each player depend on the behaviour of other players as well as on the collective strategy.  Third, having
independent detection and tracking pipelines means that the representation used for
localization is discarded, whereas re-using it would be more efficient. Finally, the
sequential approach does not scale well with the number of people in the scene, since it
requires multiple runs for a single image.

Our method aims at tackling these issues. Inspired by recent work in
multi-class object detection~\cite{Ren2015,Redmon2016} and image
labelling~\cite{Johnson2016}, we propose a single architecture that jointly
localizes multiple people, and classifies the actions of each individual as well
as their collective activity. Our model produces all the estimates in a single
forward pass and requires neither external region proposals nor pre-computed 
detections or tracking assignments. 

Our contributions can be summarized as follows:

\begin{itemize}
\setlength\itemsep{0cm}
\item We propose a unified framework for social scene understanding by simultaneously
solving three tasks in a single feed forward pass through a Neural Network: multi-person
detection, individual's action recognition, and collective activity recognition.  
Our method operates on raw image sequences and relies on joint multi-scale features 
that are shared among all the tasks. 
It allows us to fine-tune the feature extraction layers early enough to enable the model to capture the context and interactions.
\item We introduce a novel multi-object detection scheme, inspired by the classical
work on Hough transforms. Our scheme relies on probabilistic inference that jointly
refines the detection hypotheses rather than greedily discarding them, which makes
our predictions more robust.
\item We present a person-level matching Recurrent Neural Network (RNN) model to
propagate information in the temporal domain, while not having access to the the
trajectories of individuals.
\end{itemize}

In Section~\ref{sec:evaluation}, we show quantitatively that these
components contribute to the better overall performance. Our model achieves
state-of-the-art results on challenging multi-person sequences, and 
outperforms existing approaches that rely on the ground truth annotations at
test time. We demonstrate that our novel detection scheme is on par with the 
state-of-the art methods on a large-scale dataset for localizing multiple 
individuals in crowded scenes. Our implementation will be made publicly available.

\begin{figure*}[htp!]
\vspace{-0.25cm}
\begin{center}
\includegraphics[width=\textwidth]{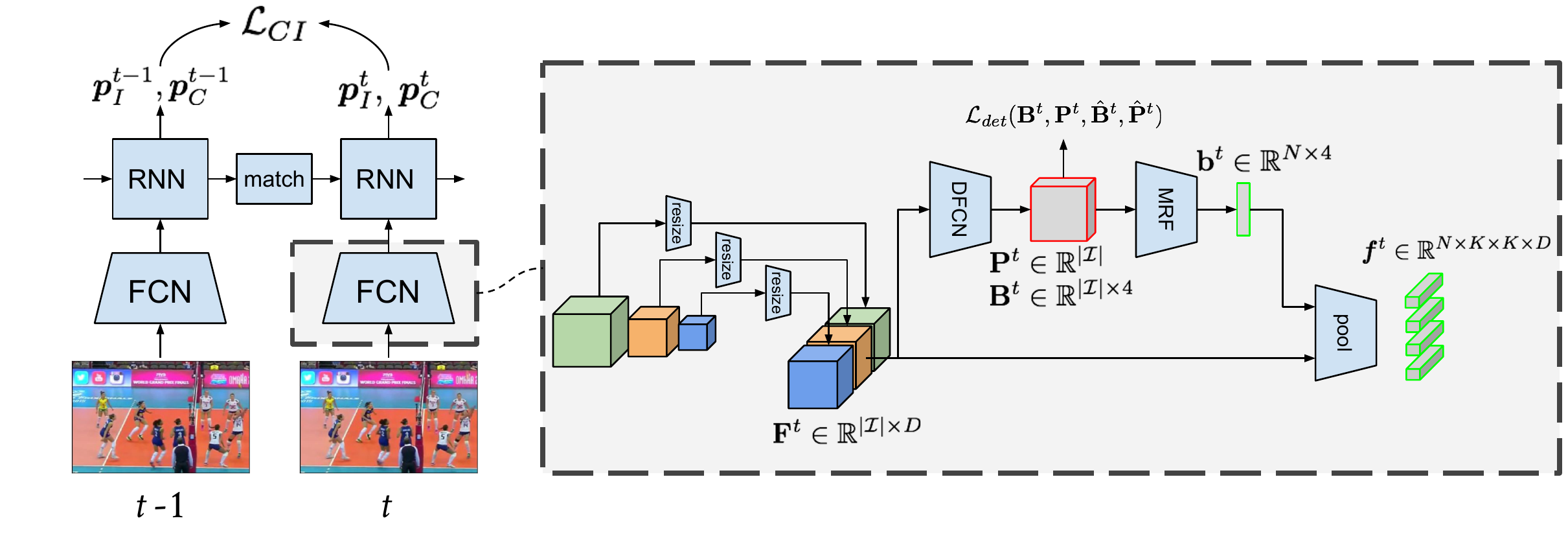}
\end{center}
\vspace{-0.5cm}
\caption{General overview of our architecture. Each frame of the given 
sequence is passed through a fully-convolutional network (FCN) to produce a 
multi-scale feature map $\bF^t$, which is then shared between the detection and action 
recognition tasks. Our detection pipeline is another fully-convolutional network (DFCN) 
that produces a dense set of detections $\bB^t$ along with the probabilities 
$\bP^t$, followed by inference in a hybrid MRF. 
The output of the MRF are reliable detections $\bb^t$ which are used to extract
fixed-sized representations $\ff^t$, which are then passed to a matching RNN that reasons 
in the temporal domain. The RNN outputs the probability of an individual's action,
$\bp_I$, and the collective activity, $\bp_c$ across time. Note that $\mL_{det}$ 
(\ref{eq:method:loss-detection}) is the
loss function for the detections, and $\mL_{CI}$ (\ref{eq:method:loss-actions}) is the loss
function for the individual and collective actions.}
\label{fi:method:overview}
\vspace{-0.5cm}
\end{figure*}
\vspace{-0.1cm}
\section{Related Work}
\vspace{-0.15cm}
\label{sec:related}
The main focus of this work is creating a unified model that can
simultaneously detect multiple individuals and recognize their
individual social actions and collective behaviour. 
In what follows, we give a short overview of the existing work on these tasks. 

\noindent \textbf{Multi-object detection} - There already exists large body of 
research in the area of object detection. Most of the current methods either 
rely on a sliding window approach~\cite{Sermanet2013,Zhang2015}, or on the object proposal
mechanism~\cite{Girshick2015,Ren2015}, followed by a CNN-based classifier. The
vast majority of those state-of-the-art methods do not reason jointly on the
presence of multiple objects, and rely on very heuristic post-processing
steps to get the final detections. A notable exception is the
ReInspect~\cite{Stewart2016} algorithm, which is specifically designed to handle
multi-object scenarios by modeling detection process in a sequential manner,
and employing a Hungarian loss to train the model end-to-end. We approach this
problem in a very different way, by doing probabilistic inference on top of 
a dense set of detection hypotheses, while also demonstrating state-of-the-art 
results on challenging crowded scenes. Another line of work that specifically 
focuses on joint multi-person detection~\cite{Fleuret2008, Bagautdinov2015} uses
generative models, however, those methods require multiple views or depth maps 
and are not applicable in monocular settings.

\noindent \textbf{Action recognition} - A large variety of methods for action
recognition traditionally rely on handcrafted features, such as
HOG~\cite{Dalal2005,Weinland10}, HOF~\cite{Laptev2008} and MBH~\cite{Wang2013}. 
More recently, data-driven approaches based on deep learning have
started to emerge, including methods based on 3D CNNs~\cite{Ji2013} and multi-stream
networks~\cite{Feichtenhofer2016,Singh2016a}. Some
methods~\cite{Wang2015,Singh2016b}, exploit the strengths of both handcrafted
features and deep-learned ones. Most of these methods rely in one way or
another on temporal cues: either through having a separate temporal
stream~\cite{Feichtenhofer2016, Singh2016b}, or directly encoding them into
compact representations~\cite{Laptev2008,Wang2013,Wang2013}. Yet another way to
handle temporal information in a data-driven way is Recurrent Neural Networks (RNNs).
Recently, it has received a lot of interest in the context of action
recognition~\cite{Singh2016a,Du2015,Veeriah2015,Donahue2015}. All these methods,
however, are focusing on recognizing actions for single individuals, and thus
are not directly applicable in multi-person settings.

\noindent \textbf{Collective activity recognition} - Historically, a large
amount of work on collective activity recognition relies on graphical models
defined on handcrafted features~\cite{Choi2014, Choi2011, Amer2014}. The
important difference of this type of methods with the single-person action
recognition approaches is that they explicitly enforce simultaneous 
reasoning on multiple people. The vast majority of the state-of-the-art methods for
recognizing multi-person activities thus also rely on some kind of structured model,
that allows sharing information between representations of individuals.
However, unlike earlier handcrafted methods, the focus of the recent
developments has shifted towards merging the discriminative power of neural networks
with structured models. In~\cite{Deng2016}, authors propose a way to refine
individual estimates obtained from CNNs through inference: they define a
trainable graphical model with nodes for all the people and the scene, and pass
messages between them to get the final scene-level
estimate. In~\cite{Ibrahim2016}, authors propose a hierarchical model that takes
into account temporal information. The model consists of two LSTMs: the first
operates on person-level representations, obtained from a CNN, which are then
max pooled and passed as input to the second LSTM capturing scene-level
representation.  \cite{Ramanathan2016} explores a slightly different
perspective: authors notice that in some settings, the activity is defined by
the actions of a single individual and propose a soft attention mechanism to
identify her. The complete model is very close to that of~\cite{Ibrahim2016},
except that the attention pooling is used instead of a max pool. All of those
methods are effective, however, they start joint reasoning in late inference
stages, thus possibly
discarding useful context information. Moreover, they all rely on ground truth
detections and/or tracks, and thus do not really solve the problem end-to-end.

Our model builds upon the existing work in that it also relies on the
discriminative power of deep learning, and employs a version of person-level
temporal model. It is also able to implicitly capture the context and perform
social scene understanding, which includes reliable localization and action
recognition, all in a single end-to-end framework.

\vspace{-0.1cm}
\section{Method}
\vspace{-0.15cm}

Our main goal is to construct comprehensive interpretations of social scenes
from raw image sequences. To this end, we propose a unified way to jointly detect multiple
interacting individuals and recognize their collective and individual actions.

\subsection{Overview}
\vspace{-0.15cm}

The general overview of our model is given in Figure~\ref{fi:method:overview}.
For every frame $\bI^t \in \mR^{H_0 \times W_0\times 3}$ in a given sequence, we first
obtain
a dense feature representation $\bF^t \in \mR^{|\mI|\times D}$, where $\mI = \{ 1,
\ldots, H \times W \}$ denotes the set of all pixel locations in the feature
map, $|\mI| = H \times W$ is the number of pixels in that map, and $D$ is the
number of features. The feature map $\bF^t$ is then shared between the
detection and action recognition tasks.
To detect, we first obtain a preliminary set of detection hypotheses,
encoded as two dense maps $\bB^t \in \mR^{|\mI|\times 4}$ and $\bP^t \in \mR^{|\mI|}$,
where at each location $i \in \mI$, $\bB^t_i$ encodes the coordinates of the bounding box,
and $\bP^t_i$ is the probability that this bounding box represents a person.
Those detections are refined jointly by inference in a hybrid Markov
Random Field (MRF). The result of the inference is a smaller set of $N$
reliable detections, encoded as bounding boxes $\bb^t \in \mR^{N \times 4}$. 
These bounding boxes are then used to smoothly extract
fixed-size representations $\ff^t_n \in \mR^{K\times K\times D}$ from the feature map
$\bF^t$, where $K$ is the size of the fixed representation in pixels. 
Representations $\ff^t_n$ are then used as inputs to the matching RNN, which 
merges the information in the temporal domain. At each time step $t$, RNN produces
probabilities $\bp^t_{I,k} \in \mR^{N_I}$ of individual actions for each detection 
$\bb^t_n$, along with the probabilities of collective activity 
$\bp^t_{C} \in \mR^{N_C}$, where
$N_I, N_c$ denote respectively the number of classes of individual and
collective actions. In the following sections, we will describe each of these
components in more detail. 


\subsection{Joint Feature Representation}
We build upon the Inception architecture~\cite{Szegedy2015} for getting our
dense feature representation, since it does not only demonstrate good performance
but is also more computationally efficient than some of the more popular
competitors~\cite{Simonyan2014, Krizhevsky2012}. 

One of the challenges when simultaneously dealing with multiple tasks
is that representations useful for one task may be quite inefficient for another. 
In our case, person detection requires
reasoning on the type of the object, whereas discriminating between actions can
require looking at lower-level details. To tackle this problem, we propose
using multi-scale features: instead of simply using the final convolutional
layer, we produce our dense feature map $\bF \in \mR^{|\mI|\times D}$ (here and later
$t$ is omitted for clarity) by concatenating multiple intermediate activation
maps. Since they do not have fitting dimensions, we resize them to the fixed
size $|\mI| = H \times W$ via differentiable bilinear interpolation. 
Note that similar approaches have been very successful for semantic
segmentation~\cite{Long2015,   Hariharan2015}, when one has to simultaneously
reason about the object class and its boundaries.


\subsection{Dense Detections}
\vspace{-0.15cm}

\begin{figure}
\begin{center}
\begin{tabular}{ccc}
\includegraphics[width=0.12\textwidth]{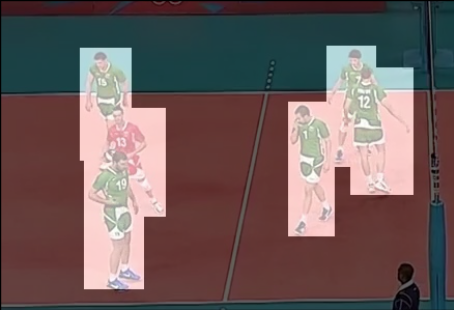} &
\includegraphics[width=0.12\textwidth]{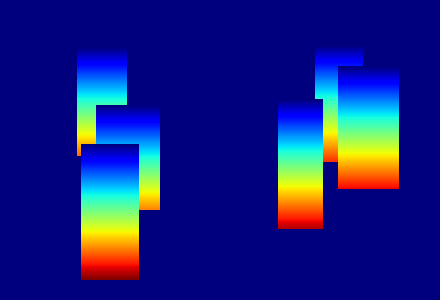} &
\includegraphics[width=0.12\textwidth]{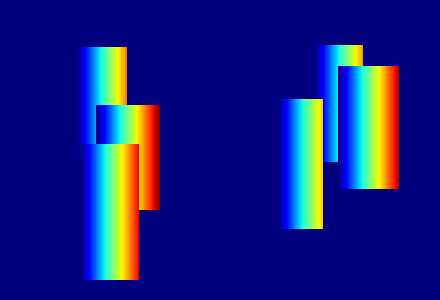} \\
\includegraphics[width=0.12\textwidth]{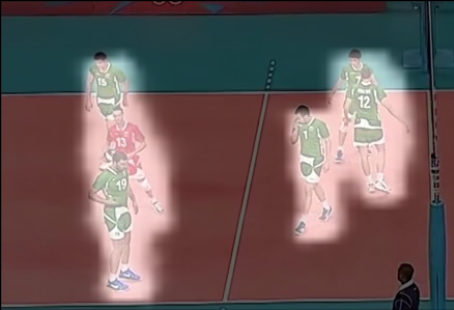} &
\includegraphics[width=0.12\textwidth]{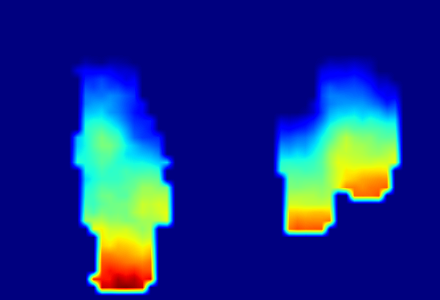} &
\includegraphics[width=0.12\textwidth]{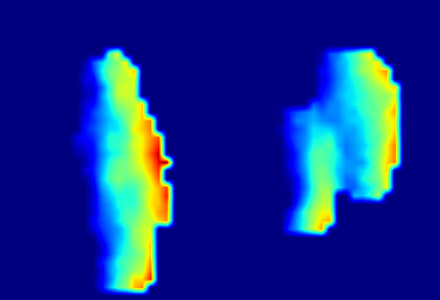} \\
\end{tabular}
\end{center}
\vspace{-0.25cm}
\caption{Example of ground truth (top) and predicted (bottom) maps. We show
  segmentation map $\bP$ projected on the original image, followed by two out of four
  channels of the regression map $\bB$, which encode respectively vertical and horizontal displacement from the location $i$ to one of the bounding box corners.}
\label{fi:method:dense-maps}
\vspace{-0.5cm}
\end{figure}

Given the output of the feature extraction stage, the goal of the detection
stage is to generate a set of reliable detections, that is, a set of
bounding box coordinates with their corresponding confidence scores. We do it in
a dense manner, meaning that, given the feature map $\bF \in \mR^{|\mI|\times D}$,
we produce two dense maps $\bB \in \mR^{|\mI|\times 4}$ and $\bP \in \mR^{|\mI|}$,
for bounding boxes coordinates and presence probability,
respectively. Essentially, $\bP$ represents a segmentation mask encoding
which parts of the image contain people, and $\bB$
represents the coordinates of the bounding boxes of the people present in the
scene, encoded relative to the pixel locations. This is illustrated by
Figure~\ref{fi:method:dense-maps}. 

We can interpret this
process of generating $\bP, \bB$ from $\bF$ in several different ways. With
respect to recent work on object
detection~\cite{Girshick2015,Redmon2016,Ren2015}, it can be seen as a
fully-convolutional network that produces a dense set of object proposals, where
each pixel of the feature map $\bF$ generates a proposal. Alternatively, we
can see this process as an advanced non-linear version of the Hough transform,
similar to Hough Forests~\cite{Gall2011, Barinova2012}. In these methods, each
patch of the image is passed through a set of decision trees, which produce a
distribution over potential object locations. The crucial differences with the
older methods are, first, leveraging deep neural network as a more powerful 
regressor and, second, the ability to use large contexts in the image, in 
particular to reason jointly about parts. 

Let us now introduce $\bB$ and $\bP$ more formally, by defining how we convert
the given ground truth object locations into dense ground truth maps
$\hat{\bB}, \hat{\bP}$. For each image $\bI$, the detection ground truth is 
given as a set of bounding boxes
$\{ (y_0,x_0,y_1,x_1)_1, \ldots, \}$. To obtain the
value for the specific location $i = (i_y, i_x) \in \mI$ of the ground truth
probability map $\hat{\bP}$, we set $\hat{\bP}_{i} = 1$ if 
$y_0 \leq i_y \leq y_1, x_0 \leq i_x \leq x_1$ for any of the ground truth boxes,
and $\hat{\bP}_{i} = 0$ otherwise. 
For the regression map, each location $i$ represents a vector 
$\hat{\bB}_{i} = (t_{y0}, t_{x0}, t_{y1}, t_{x1})$, where:

\begin{align}
\vspace{-0.5cm}
t_{y0} = (i_y - y_0) / s_y, \; t_{x0} = (i_x - x_0) / s_x \;,\\
t_{y1} = (y_1 - i_y) / s_x, \; t_{x1} = (x_1 - i_x) / s_y \;,
\label{eq:method:boxes-map}
\end{align}
where $s_y, s_x$ are scaling coefficients that are fixed, and can be taken
either as the maximum size of the bounding box over the training set, or the
size of the image. Ultimately, our formulation makes it possible to use
ground truth instance-level segmentation masks to assign each $i$ to one of the ground
truth instances. However, since these masks  are not available, and there 
can be multiple ground truth bounding boxes that contain $i$, we
assign each $i$ to the bounding box with the highest $y_0$ coordinate, 
as shown in Figure~\ref{fi:method:dense-maps}. Note that, $\hat{\bB}_i$ are only
defined only for $i: \hat{\bP}_i = 1$, and the regression loss is constructed 
accordingly.

The mapping from $\bF$ to $\bB$, $\bP$ is a fully-convolutional network,
consisting of a stack of two $3\times 3$ convolutional layers with 512 filters 
and a shortcut connection~\cite{He2016}. We use softmax activation function for
$\bP$ and ReLU for $\bB$. The loss is defined as follows:
\begin{align}
\begin{split}
\mL_{det} = & - \frac{1}{|\mI|} \sum_{i}\hat{\bP}_{i} \log \bP_{i} + \\
           & w_{reg}
\frac{1}{\sum_{i}\hat{\bP}_{i}} \cdot \sum_{i}\hat{\bP}_{i} ||\hat{\bB}_{i} -
\bB_{i}||_2^2 \;,
\label{eq:method:loss-detection}
\end{split}
\end{align}
where $w_{reg}$ is a weight that makes training focused
more on classification or regression. For datasets where classification is easy,
such as \texttt{volleyball}~\cite{Ibrahim2016}, we set it to $w_{reg} = 10$, 
whereas for cluttered scenes with large variations in appearance lower values could be
beneficial.


\subsection{Inference for Dense Detection Refinement}
\vspace{-0.15cm}
\label{sec:method:inference}

The typical approach to get the final detections given a set of proposals is to
re-score them using an additional recognition network and then run non-maxima
suppression (NMS)~\cite{Johnson2016, Ren2015}. This has several
drawbacks. First, if the amount of the proposals is large, the re-scoring stage can
be prohibitively expensive. Second, the NMS step itself is by no means optimal, and is
susceptible to greedy decisions. Instead of this commonly used technique, we
propose using a simple inference procedure that does not require re-scoring, and
makes NMS in the traditional sense unnecessary. Our key observation is that instead
of making similar hypotheses suppressing each other, one can rather make them
refine each other, thus increasing the robustness of the final estimates.

To this end, we define a hybrid MRF on top of the dense
proposal maps $\bB^*$, which we obtain by converting $\bB$ to the global image
coordinates.  For each hypothesis location $i \in \mI$ we introduce two hidden
variables, one multinomial Gaussian $\bX_i \in \mR^4$, and one categorical $A_i
\in \mI$. $\bX_i$ encodes the ``true'' coordinates of the detection, and $A_i$
encodes the assignment of the detection to one of the
hypothesis locations in $\mI$. Note that, although this assignment variable is discrete,
we  formulate our problem in a probabilistic way, through distributions, thus
allowing a detection to be ``explained'' by multiple locations. The joint
distribution over $\bX_{1:|\mI|}, A_{1:|\mI|}$ is defined as follows:

\vspace{-0.25cm}
\begin{equation}
P(\bX_{1:|\mI|}, A_{1: |\mI|}) \propto \prod_{i,j} \exp
\left(
- \frac{\ONE{[A_i = j]} \cdot ||\bX_i - \bX_j||_2^2}{2 \sigma^2} 
\right)\;,
\label{eq:model:mrf-joint}
\end{equation}
where $\sigma$ is the standard deviation parameter, which is fixed.

Intuitively, (\ref{eq:model:mrf-joint}) jointly models the relationship between 
the bounding box predictions produced by the fully-convolutional network. 
The basic assumption is that each location $i \in \mI$ on the feature map 
belongs to a single "true" detection location $j$, which can be equal to $i$, 
and the observation $\bX_i$ should not be far from the observation $\bX_j$ at this "true"
location. The goal of inference is to extract those "true" locations
and their corresponding predictions by finding the optimal assignments for $A_i$ and
values of $\bX_i$. In other words, we want to compute marginal distributions
$P(\bX_{i}), P(A_i), \forall i \in \mI$. Unfortunately, the exact integration is not
feasible, and we have to resort to an approximation. We use the mean-field approximation, 
that is, we introduce the following factorized variational distribution: 

\vspace{-0.25cm}
\begin{equation}
Q(\bX_{1:|\mI|}, A_{1:|\mI|}) =
\prod_{i} \mN(\bX_i \;; \bmu_i, \sigma^2) \cdot \mathrm{Cat}(A_i \;; \feta_i) \;,
\label{eq:model:mrf-variational}
\end{equation}
where $\bmu_i \in \mR^4$ and $\feta_i \in \mR^{|\mI|}$ are the variational
parameters of the Gaussian and categorical distributions respectively. Then,
we minimize the KL-divergence between the variational distribution
(\ref{eq:model:mrf-variational}) and the joint (\ref{eq:model:mrf-joint}), which
leads to the following fixed-point updates for the parameters of $Q(\cdot)$:

\begin{flalign}
\vspace{-0.25cm}
\eta^{\tau}_{ij} \propto -\frac{||\bmu^{\tau-1}_i - \bmu^{\tau-1}_j||^2_2}{2\sigma^2} \;,
\balpha^{\tau}_{i} =   \mathrm{softmax}(\feta^{\tau}_i) \;,
\label{eq:model:mrf-update-eta} \\
\hat{\bmu}_i^{\tau} = \sum_{j} \alpha_{ij} \bmu_j^{\tau-1} \;,
\label{eq:model:mrf-update-mu}
\vspace{-0.25cm}
\end{flalign}
where $\tau \in \{1, \ldots, \mT\}$ is the iteration number, 
$\balpha^{\tau}_{i} \in \mR^{|\mI|},
\sum_{j} \alpha^{\tau}_{ij} = 1$ is the reparameterization of $\feta^{\tau}_i$. The
complete derivation of those updates is provided in the supplementary material.

Starting from some initial $\bmu^{0}$, one can now use
(\ref{eq:model:mrf-update-eta}),
(\ref{eq:model:mrf-update-mu}) until convergence. In practice, we start with
$\bmu^{0}$ initialized from the estimates $\bB^*$, thus conditioning our model
on the observations, and only consider those $i \in \mI$, for which the
segmentation probability $\bP_{i} > \rho$, where $\rho$ is a fixed threshold. 
Furthermore, to
get $\bmu^{\tau}$ we use the following smoothed update for a fixed number of
iterations $\mT$:
\begin{equation}
\bmu^{\tau}_{i} = (1 - \lambda) \cdot \bmu^{\tau-1} + \lambda \cdot \hat{\bmu}^{\tau} \;,
\end{equation}
where $\lambda$ is a damping parameter that can be interpreted as a
step-size~\cite{Baque2016}.

To get the final set of detections, we still need to identify the most likely
hypothesis out of our final refined set $\bmu^{\mT}$. Luckily, since we also have
the estimates $\balpha^{\mT}_i$ for the assignment variables $A_i$, we can identify
them using a simple iterative scheme similar to that used in Hough
Forests~\cite{Barinova2012}. That is, we identify the hypothesis with the
largest number of locations assigned to it, then remove those
locations from consideration, and iterate until there are no unassigned
locations left. The number of assigned locations is then used as a detection
score with a very nice interpretation: a number of pixels that ``voted'' for
this detection. 

\subsection{Matching RNN for Temporal Modeling}
\vspace{-0.15cm}

Previous sections described a way to obtain a set of reliable detections
from raw images. However, temporal information is known to be a very important
feature when it comes to action recognition~\cite{Laptev2008, Wang2013}. To this
end, we propose using a matching Recurrent Neural Network, that allows us to 
merge and propagate information in the temporal domain.

For each frame $t$, given a set of $N$ detections $\bb_n^t, n \in
\{1,\ldots,N\}$, we first smoothly extract fixed-sized representations $\ff_n^t
\in \mR^{K\times K\times D}$ from the the dense feature map $\bF^t$, using
bilinear interpolation. This is in line
with the ROI-pooling~\cite{Ren2015}, widely used in object detection,  and can
be considered as a less generic version of spatial transformer
networks~\cite{Jaderberg2015}, which were also successfully used for image
captioning~\cite{Johnson2016}. Those representations $\ff^t_n$ are then passed
through a fully-connected layer, which produces more compact embeddings
$\be^t_n \in \mR^{D_e}$, where $D_e$ is the number of features in the embedded
representation. These embeddings are then used as inputs to the RNN units.

We use standard Gated Recurrent Units (GRU) ~\cite{Chung2014} for each person in 
the sequence, with a minor modification. Namely, we do not have access to the track
assignments neither during training nor testing, which means that the hidden states 
$\bh_n^t \in \mR^{D_h}$ and $\bh_n^{t+1} \in \mR^{D_h}$, where $D_h$ is the number 
of features in the hidden state, are not necessarily corresponding to the same person. 
Our solution to this is very simple: we compute the Euclidean distances between each
pair of representations at step $t$ and $t-1$, and then update the hidden state
based on those distances. A naive version that works well when the ground truth
locations are given, is to use bounding box coordinates $\bb^t, \bb^{t-1}$ as the
matching representations, and then update $\bh_n^t$ by the closest match
$\bh_{n^*}^{t-1}$:

\vspace{-0.25cm}
\begin{flalign}
n^* = \argmin_m ||\bb^{t}_n - \bb^{t-1}_m||^2_2  \;, \\
\bh^t_n = \mathrm{GRU}(\be_n^t, \bh^{t-1}_{n^*}) \;.
\label{eq:model:matching-boxes}
\end{flalign}

Alternatively, instead of bounding box coordinates $\bb^t$, one can use the
embeddings $\be^t$. This allows the model to learn a 
suitable representation, which can be potentially more robust to missing/misaligned
detections. Finally, instead of finding a \textit{single}
nearest-neighbor to make the hidden state update, we can use \textit{all} 
the previous representations, weighted by the distance in the embedding space as 
follows:
\vspace{-0.25cm}
\begin{flalign}
w^{t}_{nm} \propto \exp( - ||\be^{t}_n - \be^{t-1}_m||^2_2) \;,  
\sum_{m} w^t_{nm} = 1,\\
\hat{\bh}^{t-1} = \sum_{m} w^t_{nm} \bh_{m}^{t-1} \;,\\
\bh^t_n = \mathrm{GRU}(\be_n^t, \hat{\bh}^{t-1}) \;.
\label{eq:model:matching-embed-soft}
\end{flalign}

We experimentally evaluated all of these
matching techniques, which we call respectively \texttt{boxes},
\texttt{embed} and \texttt{embed-soft}. We provide results 
in Section~\ref{sec:evaluation}.

To get the final predictions $\bp_{C}^t$ for collective activities, we max pool over the
hidden representations $\bh^t$ followed by a softmax classifier. 
The individual actions predictions $\bp_{I,n}^t$ are computed by a separate softmax 
classifier on top of $\bh^t_n$ for each detection $n$. The loss is defined as follows:
\vspace{-0.25cm}
\begin{flalign}
\begin{split}
\mL_{CI} = &
- \frac{1}{T \cdot N_{C}}
\sum_{t,c} \hat{\bp}^t_{C,c} \log \bp^t_{C,c} \\
&- w_{I} \frac{1}{T \cdot N \cdot N_{I}} \sum_{t,n,a} \hat{\bp}^t_{I,n,a} \log
\bp^t_{I,n,a} \;,
\label{eq:method:loss-actions}
\end{split}
\end{flalign}
where $T$ is the number of frames, $N_{C}, N_{I}$ are the numbers of labels for
collective and individual actions, $N$ is the number of detections, and 
$\hat{\bp}_{*}$ is the one-hot-encoded ground truth. The weight
$w_{I}$ allows us to balance the two tasks differently, but we found that the model
is somewhat robust to the choice of this parameter. In our experiments, we set 
$w_I = 2$.

\vspace{-0.1cm}
\section{Evaluation}
\vspace{-0.15cm}
\label{sec:evaluation}

In this section, we report our results on the task of multi-person scene 
understanding and compare them to the baselines introduced in 
Section~\ref{sec:related}. We also compare our detection pipeline to 
multiple state-of-the-art detection algorithms on a challenging dataset
for multi-person detection.

\subsection{Datasets}

We evaluate our framework on the recently introduced \texttt{volleyball} 
dataset~\cite{Ibrahim2016}, since it is the only
publicly available dataset for multi-person activity recognition that is 
relatively large-scale and contains labels for people locations, as well 
as their collective and individual actions.

This dataset consists of 55 volleyball games with 4830 labelled frames, where
each player is annotated with the bounding box and one of the 9 individual actions,
and the whole scene is assigned with one of the 8 collective activity labels, which
define which part of the game is happening. For each
annotated frame, there are multiple surrounding unannotated frames available. To
get the ground truth locations of people for those, we resort to the same
appearance-based tracker as proposed by the authors of the dataset~\cite{Ibrahim2016}.




\subsection{Baselines}

We use the following baselines and versions of our approach in the evaluation:
\begin{itemize}
\setlength\itemsep{0cm}
\item \texttt{Inception-scene} - Inception-v3 network~\cite{Szegedy2015},
  pre-trained on ImageNet and fine-tuned to predict collective actions on whole
  images, without taking into account locations of individuals.
\item \texttt{Inception-person} - similar to previous baseline, but trained to
  predict individual actions based on high-resolution fixed-sized images of
  individual people, obtained from the ground truth detections.
\item \texttt{HDTM} - A 2-stage deep temporal model model~\cite{Ibrahim2016},
  consisiting of one LSTM to aggregate person-level dynamics, and one LSTM to
  aggregate scene-level temporal information. We report multiple versions of this
  baseline: the complete version which includes both scene-level and person-level 
  temporal models, \texttt{scene}, which only uses scene-level LSTM, 
  and \texttt{person}, which only uses person-level LSTM.
\item \texttt{OURS-single} - A version of our model that does not use an RNN. 
We report results for ground truth locations, as well as detections
  produced by our detection pipeline.
\item \texttt{OURS-temporal} - A complete version of our model with GRU units for
  temporal modeling. We report results both for ground truth locations and our detections,
  as well as results for different matching functions.
\end{itemize}


\subsection{Implementation Details}
\label{sec:evaluation:implementation}

All our models are trained using backpropagation using the same optimization
scheme: for all the experiments and all datasets, we use stochastic gradient descent with
ADAM~\cite{Kingma2014}, with the initial learning rate set to $10^{-5}$, and
fixed hypereparameters to $\beta_1 = 0.9, \beta_2 = 0.999, \epsilon = 10^{-8}$.

We train our model in two stages: first, we train a network on single frames, 
to jointly predict detections, individual, and collective actions. We then fix 
the weights of the feature extraction part of our model, and train our temporal RNN 
to jointly predict individual actions together with collective activities. 
Note that in fact our model is fully-differentiable, and the reason for this 
two-stage training is purely technical: backpropagation requires keeping all the
activations in memory, which is not possible for a batch of image sequences.
The total loss is simply a sum of the detection loss (\ref{eq:method:loss-detection}) 
and the action loss (\ref{eq:method:loss-actions}) for the first stage, and the action
loss for the second stage. We use a temporal window of length $T = 10$, which 
corresponds to 4 frames before the annotated frame, and 5 frames after. 

The parameters of the MRF are the same for all the experiments. We run inference 
on the bounding boxes with the probability $\bP_i$ above the threshold $\rho=0.2$, 
and set the standard deviation
$\sigma = 0.005$, step size $\lambda = 0.2$, and the number of iterations $\mT = 20$. 

Our implementation is based on TensorFlow~\cite{Abadi2015} and its running
time for a single sequence of $T = 10$ high-resolution (720x1080) images is
approximately 1.2s on a single Tesla-P100 NVIDIA GPU. 


\subsection{Multi-Person Scene Understanding}

\begin{figure*}[htp!]
\begin{center}
\begin{tabular}{ccc}
\includegraphics[width=0.31\textwidth]{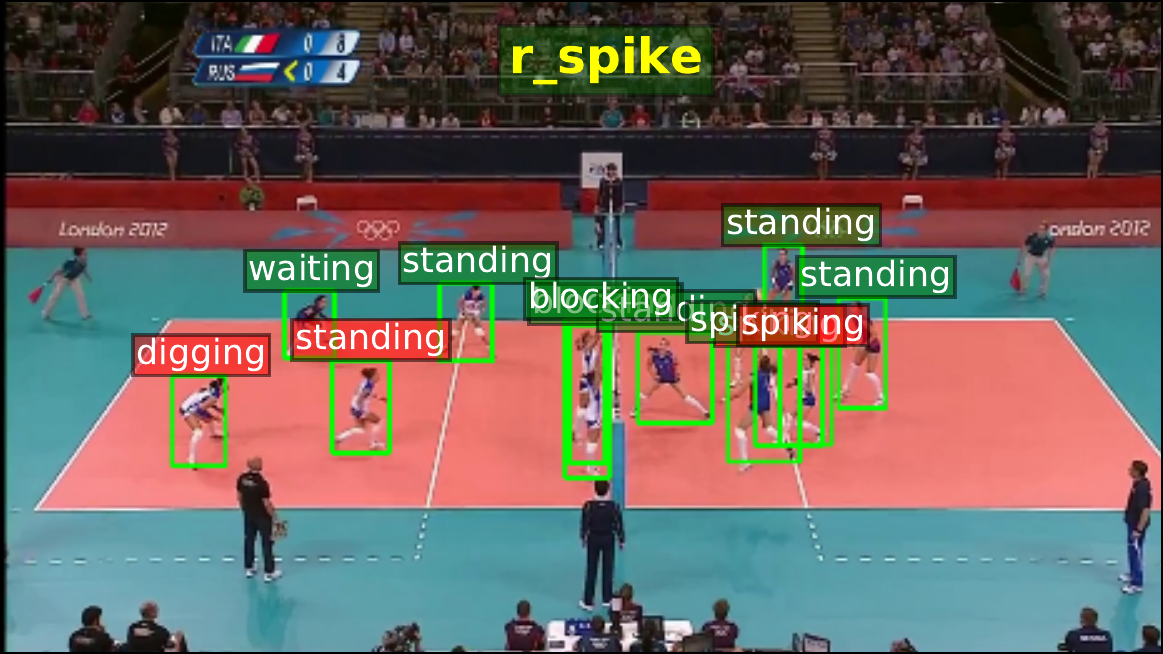} &
\includegraphics[width=0.31\textwidth]{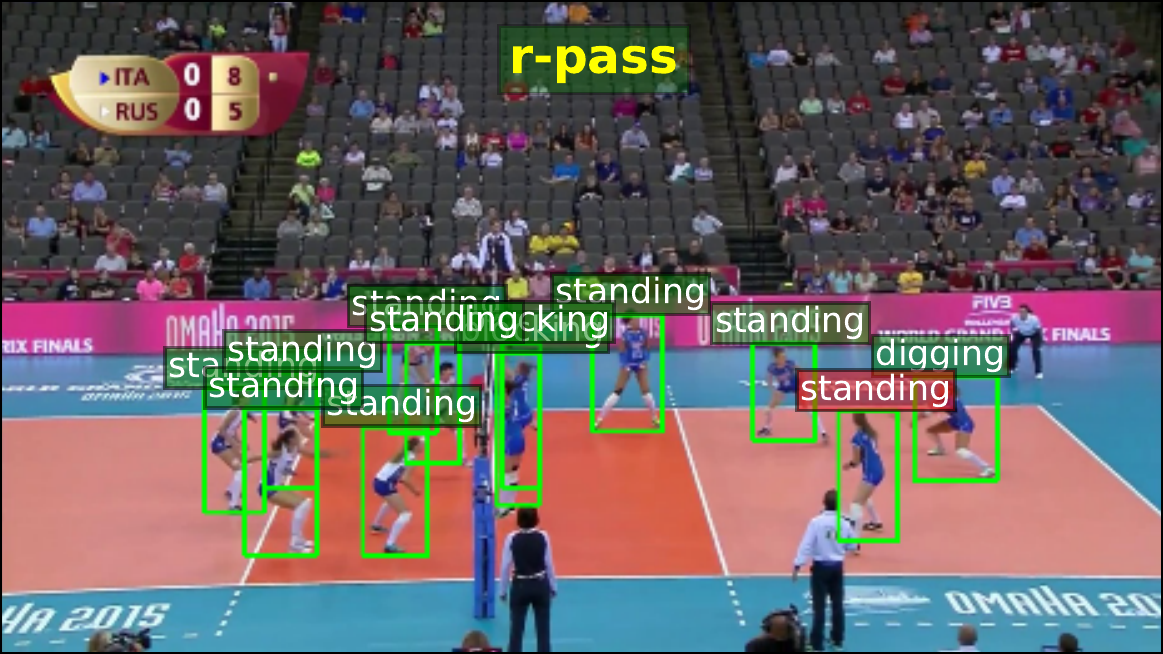} &
\includegraphics[width=0.31\textwidth]{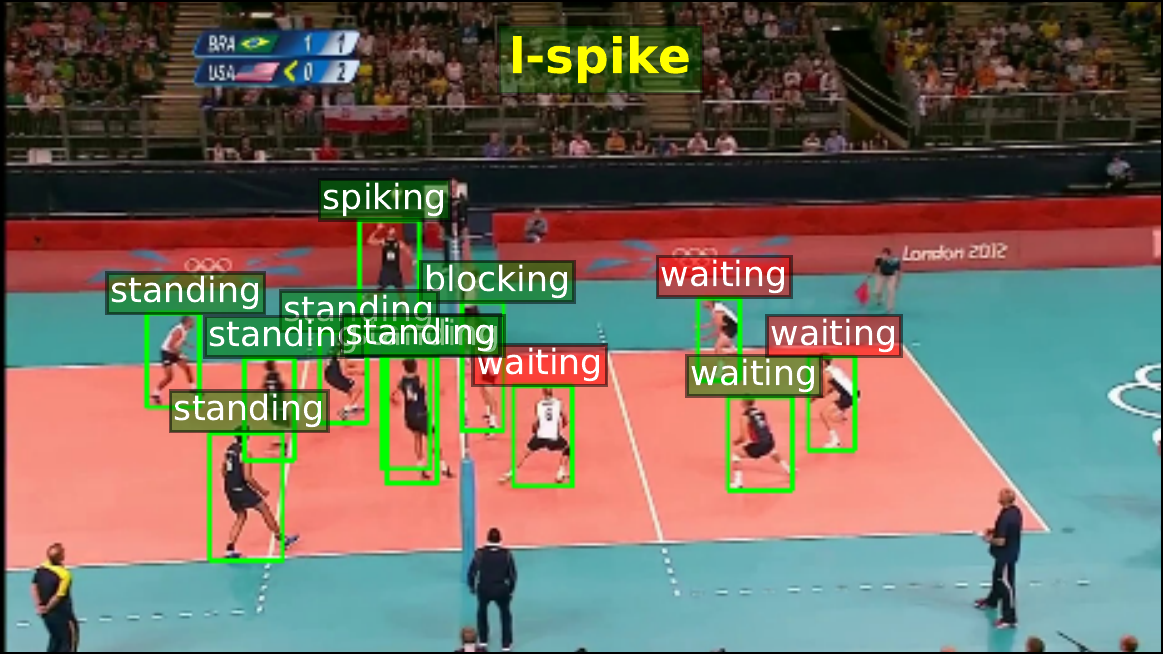} \\
\includegraphics[width=0.31\textwidth]{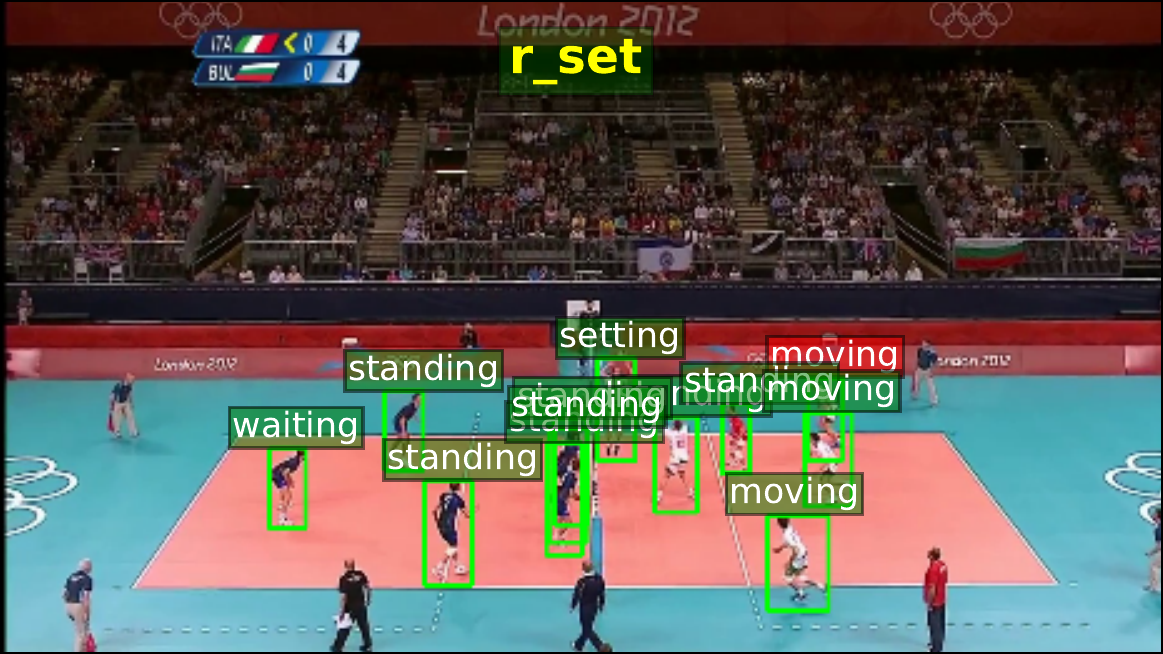} &
\includegraphics[width=0.31\textwidth]{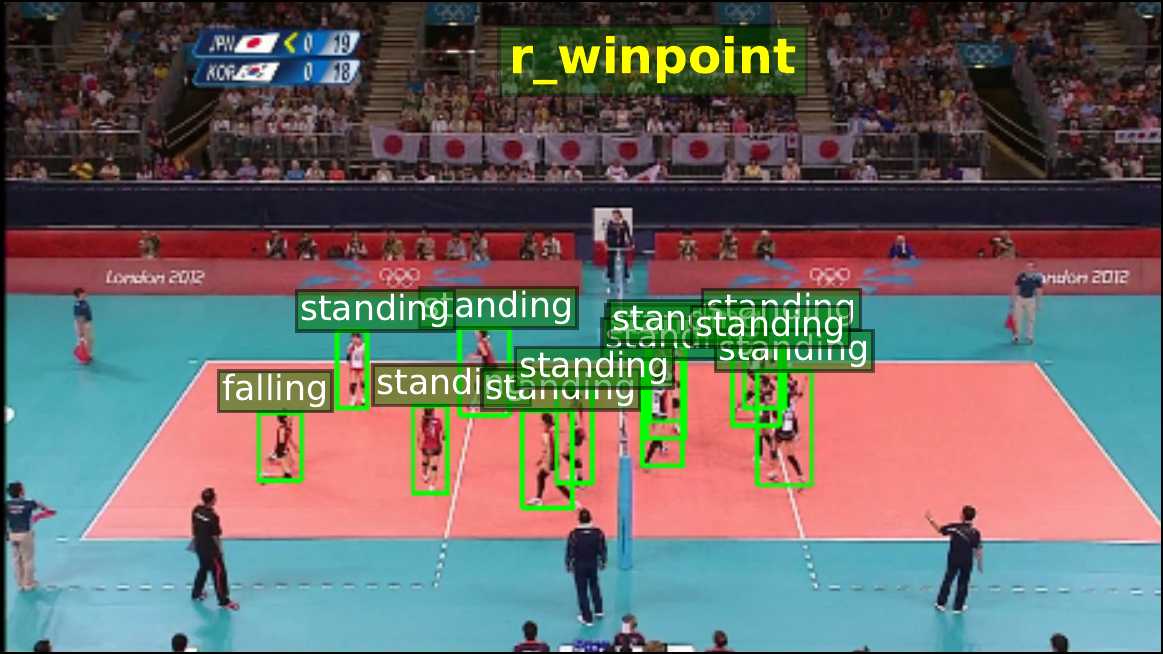} &
\includegraphics[width=0.31\textwidth]{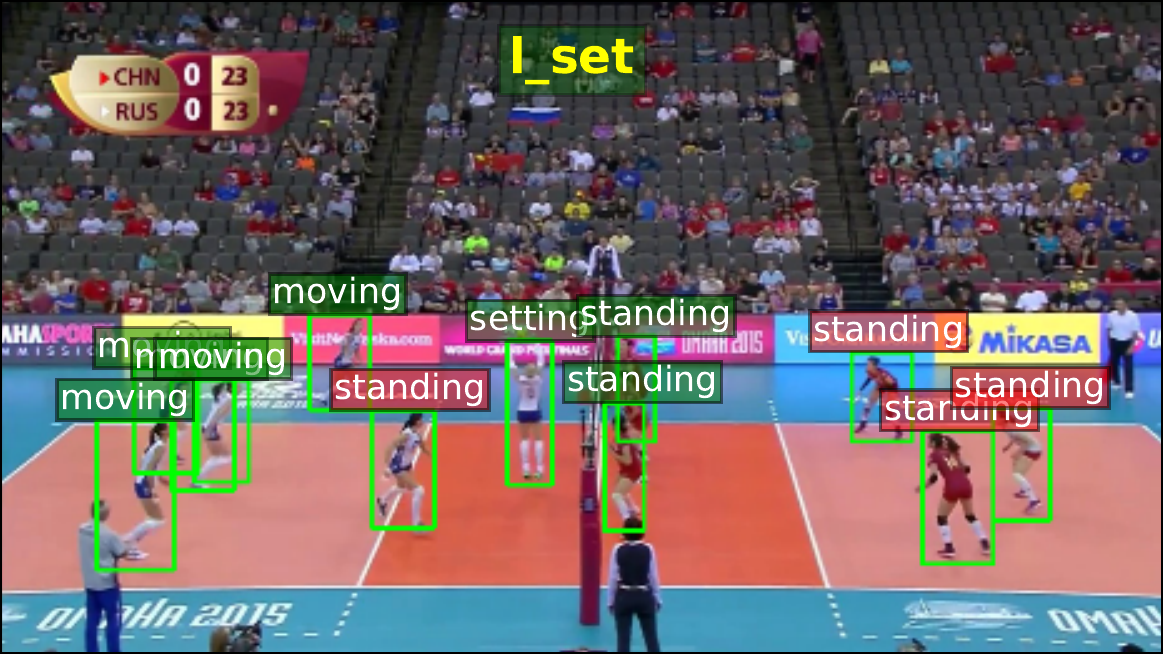} \\
\end{tabular}
\end{center}
\caption{Examples of visual results (better viewed in color). 
Green boxes around the labels correspond to correct predictions, red correspond to 
mistakes. The bounding boxes in the images are produced by our detection scheme, 
and obtained in a single pass together with the action labels.}
\label{fi:eval:visual}
\vspace{-0.25cm}
\end{figure*}

The quantitative results on the \texttt{volleyball} dataset are given in Table~\ref{tab:eval:volley-baselines}. Whenever available,
we report accuracies both for collective action recognition and individual
action recognition. For variants of our methods, we report two numbers: when the
output of our detection pipeline was used (MRF), and the ground truth bounding
boxes (GT). Our method is able to achieve state-of-the-art
performance for collective activity recognition even without ground truth
locations of the individuals and temporal reasoning. With our matching RNN,
performance improvements are even more noticeable. 
The comparison to \texttt{Inception-person}, which
was fine-tuned specifically for the single task of individual action recognition,
indicates that having a joint representation which is shared across multiple
tasks leads to an improvement in average accuracy on individual actions. When we use the
output of our detections, the drop in performance is expected, especially 
since we did not use any data augmentation to make the action recognition robust to
imperfect localization. For collective actions, having perfect localization is somewhat
less important, since the prediction is based on multiple individuals.
In Figure~\ref{fi:eval:visual} we provide some visual results, bounding boxes and 
actions labels are produced by \texttt{OURS-temporal} model with \texttt{embed-soft}
matching from raw image sequences.

\begin{table}[ht!]
\vspace{-0.25cm}
\begin{center}
\begin{tabular}{|l|c|c|}
\hline
Method                    & collective & individual \\\hline
\smttt{Inception-scene} (GT)  &  75.5      &  -       \\
\smttt{Inception-person} (GT) &   -        &  78.1    \\\hline
\smttt{HDTM-scene}~\cite{Ibrahim2016}(GT) & 74.7 & - \\
\smttt{HDTM-person}~\cite{Ibrahim2016}(GT) & 80.2  & - \\
\smttt{HDTM}~\cite{Ibrahim2016}(GT) & 81.9  & - \\\hline
\smttt{OURS-single} (MRF/GT) & 83.3 / 83.8  & 77.8 / 81.1  \\
\smttt{OURS-temporal} (MRF/GT) &  87.1 / \textbf{89.9} & 77.9 / \textbf{82.4} \\\hline
\end{tabular}
\end{center}
\caption{Results on the \texttt{volleyball} dataset. We report average accuracy
  for collective activity and individual actions. For \texttt{OURS-temporal}
  for the ground truth bounding boxes (GT) we report results with the \texttt{bbox}
  matching, and for the detections (MRF) we report results with the 
  \texttt{embed} matching.}
 \label{tab:eval:volley-baselines}
 \vspace{-0.25cm}
\end{table}

In Table~\ref{tab:eval:volley-matching} we compare different matching
strategies. For the ground truth detections, as expected, simply finding the
best match in the bounding box coordinates, \texttt{boxes}, works very well.
Interestingly, using the \texttt{embed} and \texttt{embed-soft} matching are beneficial
for the performance when detections are used instead of the ground truth. It is also
understandable: appearance is more robust than coordinates, but it also
means that our model is actually able to capture that robust appearance
representation, which might not be absolutely necessary for the prediction in a
single frame scenario. Note that, whereas for the collective actions the 
temporal data seems to help significantly, the improvement for the individual action
estimation is very modest, especially for the detections. We hypothesize that in order to
discriminate better between individual actions, it is
necessary to look at how the low-level details change, which could be potentially 
smoothed out during the spatial pooling, and thus they are hard to capture for our 
RNN. 

\renewcommand{\tabcolsep}{1pt}
\renewcommand{\tabcolsep}{6pt}



\begin{table}[ht!]
  \begin{center}
    \begin{tabular}{|l|c|c|}
      \hline
      Method  & collective & individual \\\hline
      \texttt{boxes} (MRF/GT) & 82.0 / 89.9  & 68.6 / \textbf{82.4}  \\
      \texttt{embed} (MRF/GT) & 87.1 / 90.0   & 77.9 / 81.9  \\
      \texttt{embed-soft} (MRF/GT) &  86.2 / \textbf{90.6}  & 77.4 / 81.8  \\\hline 
    \end{tabular}
  \end{center}
  \caption{Comparison of different matching strategies for the
    \texttt{volleyball} dataset. \texttt{boxes}
    corresponds  to the nearest neighbour (NN) match in the space of bounding
    box coordinates, \texttt{embed} corresponds to the NN in the embedding space
    $\be$, and \texttt{embed-soft} is a soft matching in $\be$.}
  \label{tab:eval:volley-matching}
  \vspace{-0.25cm}
\end{table}

\begin{table}[ht!]
\begin{center}
\begin{tabular}{|l|c|c|}
\hline
Method                    & collective & individual  \\\hline
\texttt{boxes} MRF  & 82.0 & 68.6  \\
\texttt{boxes} NMS  & 77.0 & 68.1  \\\hline
\texttt{embed} MRF  & \textbf{87.1} & \textbf{77.9}  \\
\texttt{embed} NMS  & 85.2 & 76.2 \\\hline
\texttt{embed-soft} MRF  & 86.2 & 77.4 \\  
\texttt{embed-soft} NMS  & 85.1 & 75.7 \\\hline
\end{tabular}
\end{center}
\caption{Comparative results of detection schemes on the \texttt{volleyball}
  dataset. We report the average accuracy for the collective and individual action
  recognition.}
\label{tab:eval:volley-detection}
\vspace{-0.15cm}
\end{table}

We also conducted experiments to see if our joint detection using MRF is
beneficial, and compare it to the traditional non-maxima
suppression, both operating on the same dense detection maps. The results for various
matching strategies are given in Table~\ref{tab:eval:volley-detection}. For all of them, 
our joint probabilistic inference leads to better accuracy than
non-maxima suppression.

\subsection{Multi-Person Detection}

For completeness, we also conducted experiments for multi-person detection using
our dense proposal network followed by a hybrid MRF. Our main competitor is the
 \texttt{ReInspect} algorithm~\cite{Stewart2016}, which was specifically designed for
 joint multi-person detection. We trained and tested our model on the \texttt{brainwash}
dataset~\cite{Stewart2016}, which contains more than 11000 training and 500 testing
images, where people are labeled by bounding boxes around their heads. 
The dataset includes some highly crowded scenes in which there are a large
number of occlusions.

Many of the bounding boxes are extremely small and thus have very little image
evidence, however, our approach allows us to simultaneously look at different
feature scales to tackle this issue. We use 5 convolutional
maps of the original Inception-v3 architecture to construct our dense
representation $\bF$. We do not tune any parameters on the validation set,
keeping them the same as for \texttt{volleyball} dataset. 

\begin{figure}[ht!]
\vspace{-0.25cm}
\begin{center}
\includegraphics[width=0.38\textwidth]{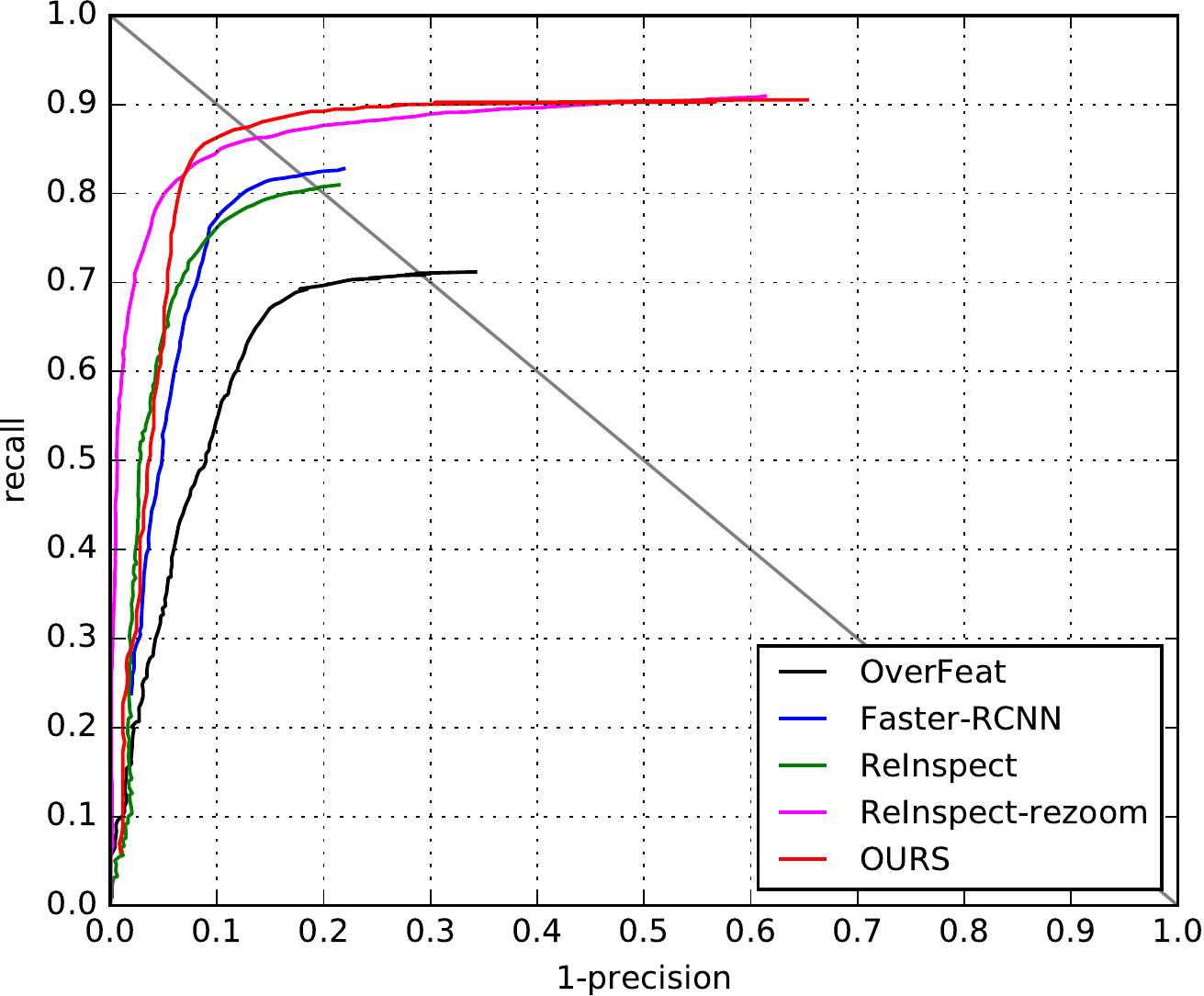}
\vspace{0.1cm}
\begin{tabular}{|l|c|c|}
\hline
Method & AP & EER \\\hline
\texttt{Overfeat}~\cite{Sermanet2013} & 0.67 & 0.71 \\
\texttt{Faster-RCNN}~\cite{Ren2015} & 0.79 & 0.80 \\
\texttt{ReInspect}~\cite{Stewart2016} & 0.78 & 0.81 \\
\texttt{ReInspect-rezoom}~\cite{Stewart2016} & \textbf{0.89} & 0.85 \\
\texttt{OURS} & 0.88 & \textbf{0.87} \\\hline
\end{tabular}
\end{center}
\caption{Results for multi-person detection on the \texttt{brainwash}~\cite{Stewart2016}
  dataset (better viewed in color). 
  Our model outperforms most of the widely used baselines, and performs on 
  par with the state-of-the-art \texttt{ReInspect-rezoom}~\cite{Stewart2016}. }
\label{fi:eval:detection-brainwash}
\end{figure}

In Figure~\ref{fi:eval:detection-brainwash} we report average precision
(AP) and equal error rate (EER)~\cite{Everingham2015}, along with the
precision-recall curves. We outperform most of the existing
detection algorithms, including widely adopted \texttt{Faster-RCNN}~\cite{Ren2015},
by a large margin, and perform very similarly to \texttt{ReInspect-rezoom}.
One of the benefits of our detection method with respect to the
\texttt{ReInspect}, is that our approach is not restricted only to detection, 
and can be also used for instance-level segmentation.


\vspace{-0.1cm}
\section{Conclusions}
\vspace{-0.15cm}
We have proposed a unified model for joint detection and activity recognition
of multiple people. Our approach does not require any external ground truth
detections nor tracks, and demonstrates state-of-the-art performance 
both on multi-person scene understanding and detection datasets. 
Future work will apply the proposed framework to explicitly capture 
and understand human interactions.

\newpage
{\small
\bibliographystyle{ieee}
\bibliography{egbib}

\begin{thebibliography}{10}\itemsep=-1pt

\bibitem{Abadi2015}
M.~Abadi, A.~Agarwal, P.~Barham, E.~Brevdo, Z.~Chen, C.~Citro, G.~S. Corrado,
  A.~Davis, J.~Dean, M.~Devin, et~al.
\newblock Tensorflow: Large-scale machine learning on heterogeneous systems,
  2015.
\newblock {\em Software available from tensorflow. org}, 1, 2015.

\bibitem{Amer2014}
M.~R. Amer, P.~Lei, and S.~Todorovic.
\newblock Hirf: Hierarchical random field for collective activity recognition
  in videos.
\newblock In {\em European Conference on Computer Vision}, pages 572--585.
  Springer, 2014.

\bibitem{Bagautdinov2015}
T.~Bagautdinov, F.~Fleuret, and P.~Fua.
\newblock Probability occupancy maps for occluded depth images.
\newblock In {\em Proceedings of the IEEE Conference on Computer Vision and
  Pattern Recognition}, pages 2829--2837, 2015.

\bibitem{Baque2016}
P.~Baque, T.~Bagautdinov, F.~Fleuret, and P.~Fua.
\newblock Principled parallel mean-field inference for discrete random fields.
\newblock In {\em The IEEE Conference on Computer Vision and Pattern
  Recognition (CVPR)}, June 2016.

\bibitem{Barinova2012}
O.~Barinova, V.~Lempitsky, and P.~Kholi.
\newblock On detection of multiple object instances using hough transforms.
\newblock {\em IEEE Transactions on Pattern Analysis and Machine Intelligence},
  34(9):1773--1784, 2012.

\bibitem{Choi2014}
W.~Choi and S.~Savarese.
\newblock Understanding collective activitiesof people from videos.
\newblock {\em IEEE transactions on pattern analysis and machine intelligence},
  36(6):1242--1257, 2014.

\bibitem{Choi2011}
W.~Choi, K.~Shahid, and S.~Savarese.
\newblock Learning context for collective activity recognition.
\newblock In {\em Computer Vision and Pattern Recognition (CVPR), 2011 IEEE
  Conference on}, pages 3273--3280. IEEE, 2011.

\bibitem{Chung2014}
J.~Chung, C.~Gulcehre, K.~Cho, and Y.~Bengio.
\newblock Empirical evaluation of gated recurrent neural networks on sequence
  modeling.
\newblock {\em arXiv preprint arXiv:1412.3555}, 2014.

\bibitem{Dalal2005}
N.~Dalal and B.~Triggs.
\newblock Histograms of oriented gradients for human detection.
\newblock In {\em 2005 IEEE Computer Society Conference on Computer Vision and
  Pattern Recognition (CVPR'05)}, volume~1, pages 886--893. IEEE, 2005.

\bibitem{Deng2016}
Z.~Deng, A.~Vahdat, H.~Hu, and G.~Mori.
\newblock Structure inference machines: Recurrent neural networks for analyzing
  relations in group activity recognition.
\newblock In {\em The IEEE Conference on Computer Vision and Pattern
  Recognition (CVPR)}, June 2016.

\bibitem{Donahue2015}
J.~Donahue, L.~Anne~Hendricks, S.~Guadarrama, M.~Rohrbach, S.~Venugopalan,
  K.~Saenko, and T.~Darrell.
\newblock Long-term recurrent convolutional networks for visual recognition and
  description.
\newblock In {\em Proceedings of the IEEE Conference on Computer Vision and
  Pattern Recognition}, pages 2625--2634, 2015.

\bibitem{Du2015}
Y.~Du, W.~Wang, and L.~Wang.
\newblock Hierarchical recurrent neural network for skeleton based action
  recognition.
\newblock In {\em Proceedings of the IEEE Conference on Computer Vision and
  Pattern Recognition}, pages 1110--1118, 2015.

\bibitem{Everingham2015}
M.~Everingham, S.~A. Eslami, L.~Van~Gool, C.~K. Williams, J.~Winn, and
  A.~Zisserman.
\newblock The pascal visual object classes challenge: A retrospective.
\newblock {\em International Journal of Computer Vision}, 111(1):98--136, 2015.

\bibitem{Feichtenhofer2016}
C.~Feichtenhofer, A.~Pinz, and A.~Zisserman.
\newblock Convolutional two-stream network fusion for video action recognition.
\newblock In {\em The IEEE Conference on Computer Vision and Pattern
  Recognition (CVPR)}, June 2016.

\bibitem{Fleuret2008}
F.~Fleuret, J.~Berclaz, R.~Lengagne, and P.~Fua.
\newblock Multicamera people tracking with a probabilistic occupancy map.
\newblock {\em IEEE Transactions on Pattern Analysis and Machine Intelligence},
  30(2):267--282, 2008.

\bibitem{Gall2011}
J.~Gall, A.~Yao, N.~Razavi, L.~Van~Gool, and V.~Lempitsky.
\newblock Hough forests for object detection, tracking, and action recognition.
\newblock {\em IEEE transactions on pattern analysis and machine intelligence},
  33(11):2188--2202, 2011.

\bibitem{Girshick2015}
R.~Girshick.
\newblock Fast r-cnn.
\newblock In {\em Proceedings of the IEEE International Conference on Computer
  Vision}, pages 1440--1448, 2015.

\bibitem{Hariharan2015}
B.~Hariharan, P.~Arbel{\'a}ez, R.~Girshick, and J.~Malik.
\newblock Hypercolumns for object segmentation and fine-grained localization.
\newblock In {\em Proceedings of the IEEE Conference on Computer Vision and
  Pattern Recognition}, pages 447--456, 2015.

\bibitem{He2016}
K.~He, X.~Zhang, S.~Ren, and J.~Sun.
\newblock Deep residual learning for image recognition.
\newblock In {\em The IEEE Conference on Computer Vision and Pattern
  Recognition (CVPR)}, June 2016.

\bibitem{Ibrahim2016}
M.~S. Ibrahim, S.~Muralidharan, Z.~Deng, A.~Vahdat, and G.~Mori.
\newblock A hierarchical deep temporal model for group activity recognition.
\newblock In {\em The IEEE Conference on Computer Vision and Pattern
  Recognition (CVPR)}, June 2016.

\bibitem{Jaderberg2015}
M.~Jaderberg, K.~Simonyan, A.~Zisserman, et~al.
\newblock Spatial transformer networks.
\newblock In {\em Advances in Neural Information Processing Systems}, pages
  2017--2025, 2015.

\bibitem{Ji2013}
S.~Ji, W.~Xu, M.~Yang, and K.~Yu.
\newblock 3d convolutional neural networks for human action recognition.
\newblock {\em IEEE transactions on pattern analysis and machine intelligence},
  35(1):221--231, 2013.

\bibitem{Johnson2016}
J.~Johnson, A.~Karpathy, and L.~Fei-Fei.
\newblock Densecap: Fully convolutional localization networks for dense
  captioning.
\newblock In {\em The IEEE Conference on Computer Vision and Pattern
  Recognition (CVPR)}, June 2016.

\bibitem{Kingma2014}
D.~Kingma and J.~Ba.
\newblock Adam: A method for stochastic optimization.
\newblock {\em arXiv preprint arXiv:1412.6980}, 2014.

\bibitem{Krizhevsky2012}
A.~Krizhevsky, I.~Sutskever, and G.~E. Hinton.
\newblock Imagenet classification with deep convolutional neural networks.
\newblock In {\em Advances in neural information processing systems}, pages
  1097--1105, 2012.

\bibitem{Laptev2008}
I.~Laptev, M.~Marszalek, C.~Schmid, and B.~Rozenfeld.
\newblock Learning realistic human actions from movies.
\newblock In {\em Computer Vision and Pattern Recognition, 2008. CVPR 2008.
  IEEE Conference on}, pages 1--8. IEEE, 2008.

\bibitem{Long2015}
J.~Long, E.~Shelhamer, and T.~Darrell.
\newblock Fully convolutional networks for semantic segmentation.
\newblock In {\em Proceedings of the IEEE Conference on Computer Vision and
  Pattern Recognition}, pages 3431--3440, 2015.

\bibitem{Ramanathan2016}
V.~Ramanathan, J.~Huang, S.~Abu-El-Haija, A.~Gorban, K.~Murphy, and L.~Fei-Fei.
\newblock Detecting events and key actors in multi-person videos.
\newblock In {\em The IEEE Conference on Computer Vision and Pattern
  Recognition (CVPR)}, June 2016.

\bibitem{Redmon2016}
J.~Redmon, S.~Divvala, R.~Girshick, and A.~Farhadi.
\newblock You only look once: Unified, real-time object detection.
\newblock In {\em The IEEE Conference on Computer Vision and Pattern
  Recognition (CVPR)}, June 2016.

\bibitem{Ren2015}
S.~Ren, K.~He, R.~Girshick, and J.~Sun.
\newblock Faster r-cnn: Towards real-time object detection with region proposal
  networks.
\newblock In {\em Advances in neural information processing systems}, pages
  91--99, 2015.

\bibitem{Sermanet2013}
P.~Sermanet, D.~Eigen, X.~Zhang, M.~Mathieu, R.~Fergus, and Y.~LeCun.
\newblock Overfeat: Integrated recognition, localization and detection using
  convolutional networks.
\newblock {\em arXiv preprint arXiv:1312.6229}, 2013.

\bibitem{Simonyan2014}
K.~Simonyan and A.~Zisserman.
\newblock Very deep convolutional networks for large-scale image recognition.
\newblock {\em arXiv preprint arXiv:1409.1556}, 2014.

\bibitem{Singh2016a}
B.~Singh, T.~K. Marks, M.~Jones, O.~Tuzel, and M.~Shao.
\newblock A multi-stream bi-directional recurrent neural network for
  fine-grained action detection.
\newblock In {\em The IEEE Conference on Computer Vision and Pattern
  Recognition (CVPR)}, June 2016.

\bibitem{Singh2016b}
S.~Singh, C.~Arora, and C.~V. Jawahar.
\newblock First person action recognition using deep learned descriptors.
\newblock In {\em The IEEE Conference on Computer Vision and Pattern
  Recognition (CVPR)}, June 2016.

\bibitem{Stewart2016}
R.~Stewart, M.~Andriluka, and A.~Y. Ng.
\newblock End-to-end people detection in crowded scenes.
\newblock In {\em The IEEE Conference on Computer Vision and Pattern
  Recognition (CVPR)}, June 2016.

\bibitem{Szegedy2015}
C.~Szegedy, V.~Vanhoucke, S.~Ioffe, J.~Shlens, and Z.~Wojna.
\newblock Rethinking the inception architecture for computer vision.
\newblock {\em arXiv preprint arXiv:1512.00567}, 2015.

\bibitem{Veeriah2015}
V.~Veeriah, N.~Zhuang, and G.-J. Qi.
\newblock Differential recurrent neural networks for action recognition.
\newblock In {\em Proceedings of the IEEE International Conference on Computer
  Vision}, pages 4041--4049, 2015.

\bibitem{Wang2013}
H.~Wang, A.~Kl{\"a}ser, C.~Schmid, and C.-L. Liu.
\newblock Dense trajectories and motion boundary descriptors for action
  recognition.
\newblock {\em International journal of computer vision}, 103(1):60--79, 2013.

\bibitem{Wang2015}
L.~Wang, Y.~Qiao, and X.~Tang.
\newblock Action recognition with trajectory-pooled deep-convolutional
  descriptors.
\newblock In {\em Proceedings of the IEEE Conference on Computer Vision and
  Pattern Recognition}, pages 4305--4314, 2015.

\bibitem{Weinland10}
D.~Weinland, M.~Ozuysal, and P.~Fua.
\newblock {Making Action Recognition Robust to Occlusions and Viewpoint
  Changes}.
\newblock 2010.

\bibitem{Zhang2015}
S.~Zhang, R.~Benenson, and B.~Schiele.
\newblock Filtered feature channels for pedestrian detection.
\newblock In {\em Proceedings of the IEEE Conference on Computer Vision and
  Pattern Recognition}, pages 1751--1760, 2015.

\end{thebibliography}
}

\end{document}